\documentclass{article}
\usepackage{spconf, amsmath, graphicx, hyperref}
\usepackage{tabularx}
\usepackage{graphicx} 
\usepackage{adjustbox}
\usepackage{amsfonts}
\usepackage{booktabs}
\usepackage{subfigure}
\usepackage{algorithm}
\usepackage{algpseudocode}
\usepackage{svg}
\usepackage[table]{xcolor}
\usepackage{bbding}
\usepackage{multirow}

\title{CARE: Multi-Task Pretraining for Latent Continuous Action Representation in Robot Control}
\name{Jiaqi Shi$^{1, 2\ddagger}$\qquad Xulong Zhang$^{1\ddagger}$\qquad Xiaoyang Qu$^{1}$\qquad Jianzong Wang$^{1\dagger}$\thanks{$^\dagger$Corresponding author is Jianzong Wang (jzwang@188.com).}\thanks{$^\ddagger$ Both authors have equal contributions.}\thanks{This work is supported by Shenzhen-Hong Kong Joint Funding Project (Category A) under grant No. SGDX20240115103359001.}}
\address{$^1$Ping An Technology (Shenzhen) Co., Ltd., Shenzhen, China, \\
$^2$University of Science and Technology of China, Hefei, China}
\begin{document}
\ninept
\maketitle
\begin{abstract}
Recent advances in Vision-Language-Action (VLA) models have shown promise for robot control, but their dependence on action supervision limits scalability and generalization. To address this challenge, we introduce CARE, a novel framework designed to train VLA models for robotic task execution. Unlike existing methods that depend on action annotations during pretraining, CARE eliminates the need for explicit action labels by leveraging only video-text pairs. These weakly aligned data sources enable the model to learn continuous latent action representations through a newly designed multi-task pretraining objective. During fine-tuning, a small set of labeled data is used to train the action head for control. Experimental results across various simulation tasks demonstrate CARE’s superior success rate, semantic interpretability, and ability to avoid shortcut learning. These results underscore CARE’s scalability, interpretability, and effectiveness in robotic control with weak supervision. 
\end{abstract}
\section{Introduction}
\label{sec:intro}
\begin{figure*}[t]
    \centering
    \includegraphics[width=0.9
    \linewidth]{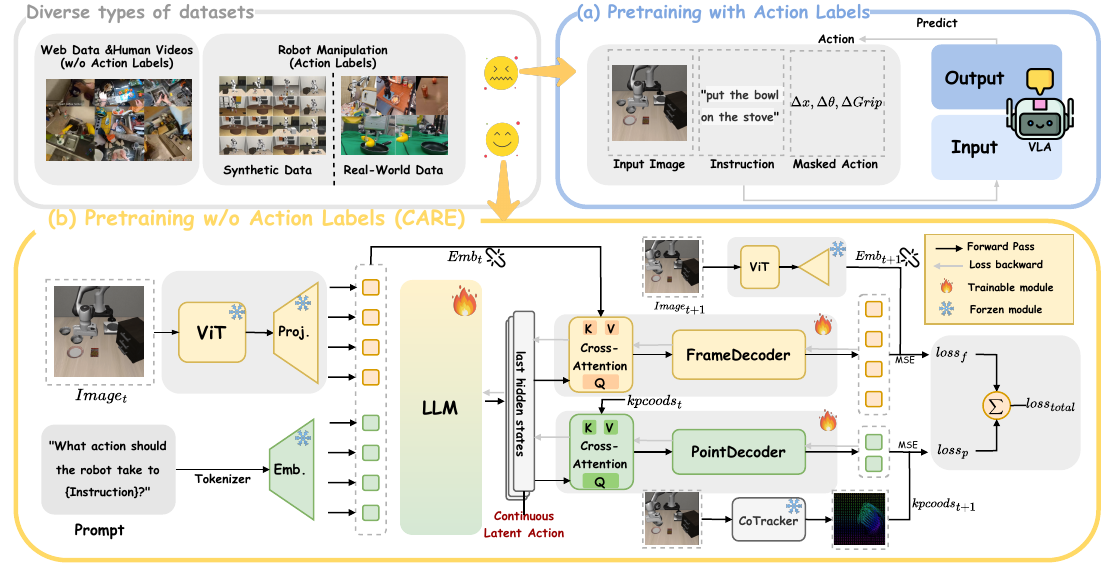}
    \\[-2.5ex]
    \caption{\textbf{Overview of CARE. } Compared to (a) the VLA training approach that requires a large number of action labels, (b) our proposed multitask learning-based pre-training method, CARE, uses only instruction-image pairs as input. Without relying on action labels, it still enables the VLM to predict potential actions. }
    \label{fig:overview}
    \vspace{-2.5ex}
\end{figure*}
Recent advancements in Visual-Language-Action (VLA) models show significant potential in robotic control by integrating visual perception, natural language understanding, and motion control to perform tasks like grasping, manipulation, and tool use \cite{brohan2022rt, driess2023palm, bjorck2025gr00t,nav,jiaziqi,jiaziqi2,aaai,pointaction}. While these models effectively bridge the gap between high-level language instructions and low-level execution, their development is heavily constrained by the costly and time-consuming nature of large-scale action supervision during pretraining \cite{o2024open}. Obtaining precise action annotations, such as joint angles or end-effector trajectories, is often impractical, especially when scaling to diverse robotic morphologies or transferring datasets across domains. 

A promising direction for scalable robotic development involves leveraging abundant, action-unlabeled video data from diverse sources. The Genie model \cite{bruce2024genie} introduced a scalable Latent Action Model (LAM) that enables generative models to create interactive environments. Using an encoder-decoder architecture with a small discrete codebook, Genie extracts discrete latent actions, providing a foundation for subsequent work. LAPA \cite{ye2024latent} first learns discrete latent actions between image frames using a Vector Quantized Variational Autoencoder (VQ-VAE) objective \cite{van2017neural}. However, the pixel-level reconstruction objective in LAM often leads models to focus on static image features, such as background textures or color distributions, rather than action-driven changes \cite{mccarthy2024towards}. The Moto model \cite{chen2412moto} addressed this by shifting the training objective from next-frame reconstruction to predicting high-level semantic embeddings using a Vision Transformer (ViT) encoder. Nevertheless, these approaches represent latent actions as discrete codebooks. The fixed size of discrete codebooks limits representation capacity, potentially failing to capture fine-grained variations in continuous action spaces. To address this, the COMO model \cite{yang2025learning} proposed continuous latent action representations, replacing the codebook with a cross-attention mechanism to capture more granular latent action representations. 

Despite these advancements, three challenges remain in latent action model research. First, bias propagation persists. Quantization errors and codebook collapse in VQ-VAE-based LAMs are fully inherited by VLA models, which cannot correct these errors. Second, latent action representations lack explicit action encoding. \cite{mccarthy2024towards} noted that, as the training objective focuses on next-frame reconstruction, inferred latent actions merely compress differences between consecutive frames, even when control actions are not the cause of these differences. Third, models risk "shortcut learning". The single training objective and LAM structure can lead models to degenerate into future frame predictors rather than latent action modelers \cite{schmidt2023learning}. 

To address these issues, we propose a novel pretraining strategy named CARE. We integrate LAM-training into the pre-training process of Vision Language Models (VLMs). Specifically, we replace the LAM encoder with a VLM and add multiple decoders for different tasks, forming an encoder-multi-decoder architecture. To address the second and third issues, we employ multi-task learning to enhance the explicit representation of latent actions. By guiding the VLM to generate latent actions through multi-task training and validating these representations with multiple decoders, we improve action encoding. Unlike prior work relying on next-frame reconstruction, we introduce keypoint trajectory prediction as a training objective. This allows the model to reconstruct the next frame while predicting keypoint trajectory changes, emphasizing action-driven positional changes. Experiments demonstrate that this multi-task learning approach enhances explicit action representation and prevents "shortcut learning". Our contributions are as follows:
\begin{itemize}
    \item We propose CARE, an unsupervised pretraining strategy that seamlessly integrates LAM training into the VLM pretraining pipeline, reducing the overall VLA training process from four stages to three. 
    \item We introduce a multi-task learning approach with keypoint trajectory prediction to enrich latent action representations, thereby improving explicit action encoding and mitigating shortcut learning. 
    \item We demonstrate through extensive simulations that our pretraining strategy provides significant advantages over prior label-free approaches while remaining competitive with methods that rely on action labels. 
\end{itemize}
\vspace{-3ex}
\section{METHODOLOGY}
\begin{figure*}[t]
    \centering
    \includegraphics[width=0.9\linewidth]{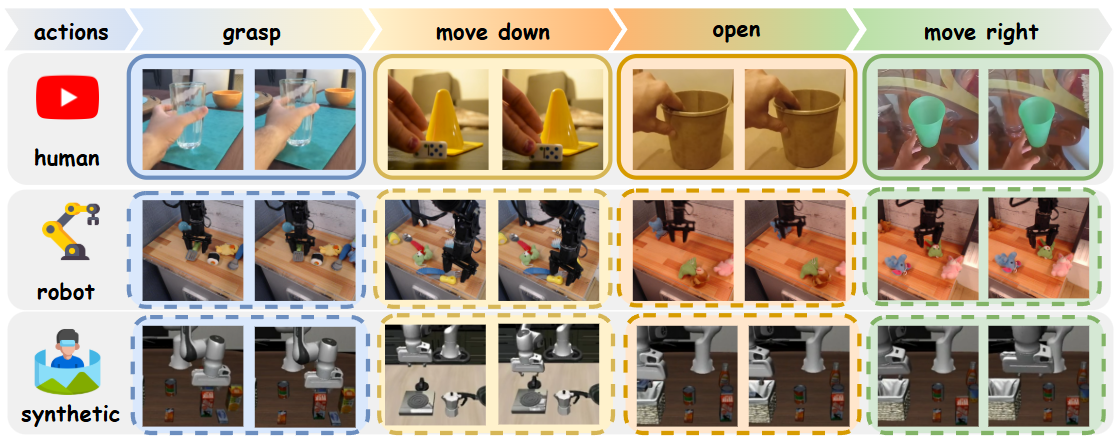}
    \\[-2ex]
    \caption{\textbf{Datasets of pre-training stage and post-training stage. }}
    \label{fig:dataset}
    \vspace{-2ex}
\end{figure*}
This section introduces our CARE pretraining framework for learning a latent vision-language model without action labels (see Fig. \ref{fig:overview}). The training process comprises two stages: (1) unsupervised pretraining of the latent VLM, and (2) supervised fine-tuning with a small-scale robot dataset.

\vspace{-1.5ex}
\subsection{Latent Vision-Language Model}
Our Latent Vision-Language Model (VLM) architecture is built upon the Prismatic-7B VLM \cite{karamcheti2024prismatic}, with a core structure consisting of a visual encoder, a projector, and a 7B-parameter Llama Large Language Model (LLM) backbone. In the pretraining stage, we fine-tune this backbone to learn latent continuous action representations. 

We define the latent action prediction problem as a “vision-language” task, where the input consists of an observation image and a natural language task instruction. Formally, a VLM takes as input an image $x_{img}$ and a prompt $p$. The model receives latent-action placeholders appended to $p$ so it can predict the action dimensions. Then, we put $p$ to the tokenizer where the llm backbone uses to get the prompt tokens $x_{ids}$. The text embedding encoder takes $x_{ids}$ as input and outputs text prompt embeddings $f_{T}$. 

The visual encoder is composed of two pretrained models, SigLIP\cite{zhai2023sigmoid} and DinoV2\cite{oquab2023dinov2}. The visual encoder processes $x_{img}$ to produce two feature sequences $f_{sig}\in \mathbb{R}^{N_{p}\times D_{v}}$ and $f_{dino}\in \mathbb{R}^{N_{p}\times D_{v}}$. These features are concatenated into $f_{cat}\in \mathbb{R}^{N_{p}\times 2D_{v}}$ and then projected by a two-layer MLP $Proj_{\zeta}(\cdot)\in \mathbb{R}^{2D_{v}\to D_{l}}$ to obtain the visual features $f_{v}$. Finally, we concatenate the text features $f_{T}$ and $f_{v}$ to form the LLM's full input. The final hidden states $h_{last}=LM_{\theta}([f_T;f_v])$ are generated, and the part corresponding to the action dimensions is used as the continuous latent action representation $z$ for multi-task learning. 

\vspace{-1.5ex}
\subsection{Continuous Latent Action with Multi-task Learning}
Multi-task learning involves the problem of optimizing a model with respect to multiple objectives. In our work, we apply multitask learning to generate latent continuous action representations. 

\noindent \textbf{Multi-Task Definition. }
 We define two training objectives. The first training objective is similar to the one used in previous work for training LAM (\cite{ye2024latent, bruce2024genie, chen2412moto}), which involves inputting the vision feature $f_{v}^{t}\in \mathbb{R}^{B\times N_{p}\times D_{l}}$ of the current frame and the latent action representation $z$ to predict the vision feature $f_{v}^{t+1}$ of the next frame. 
 
However, the single task of predicting the next frame, as discussed earlier, has certain issues. Therefore, we introduce a new training objective - point track prediction. In the point track prediction task, the input $k_{t}\in \mathbb{R}^{B\times256\times2}$ consists of the two-dimensional coordinates($x$ and $y$) of evenly distributed points in the current frame, with a shape of $256\times 2$. The two-dimensional coordinates of the current frame points $k_{t}$ and the latent action representation $z_{t}$ as inputs to predict the two-dimensional coordinates of the points in the next frame $k_{t+1}$. Specially, we use Co-Tracker\cite{karaev23cotracker}, a pre-trained model, to acquire $k_t$ and $k_{t+1}$. 

For the frame prediction task, we use the cross-attention mechanism to compute the relationship $z_{f}$ between current frame feature and latent action representation as $z_{f}=softmax(\frac{QK^{T}}{\sqrt{d}})V$
, where $Q=z_{t}W_{z}+b_{z}$ represents the query projection, and $K, V=f_{v}^{t}W_{f}+b_{f}$ represent the key and value projections, respectively. Then, we input $z_{f}$ into the frame decoder to predict the feature vector $f^{t+1}_{v}$ of the next frame. For the point track prediction task, we similarly first fuse $k_{t}$ and $z_{t}$ using cross-attention to obtain $z_{k}$. Then, we input $z_{k}$ into the point decoder to predict the point coordinates $k_{t+1}$ of the next frame. 

\noindent \textbf{Loss Function Definition. }
Given a video-text pair dataset $\mathcal{D}=\{(x_t, p_t)_{t=1}^{T_{i}}\}_{i=1}^{\mathcal{N}}$, where $\mathcal{N}$ represents the number of demonstrations. We aim to pretrain a latent policy $\pi_{\phi}:\mathcal{X}\times\mathcal{P}\to\mathcal{Z}$ with multi-task learning. The loss of frame prediction task could be described as $\mathcal{L}_{f}^{t}=MSE(f_{v}^{t+1}, \hat{f}_{v}^{t+1})$. 
The loss of point track prediction could be described as $\mathcal{L}_{p}^{t}=MSE(k^{t+1}, \hat{k}^{t+1})$. 
We adopt the Uncertainty-Weighted Loss (UWL) framework of \cite{kendall2018multi} to combine the objectives in our multi-task training. This approach is derived by maximizing a Gaussian likelihood under the assumption of task-independent uncertainty. For Latent VLM, the model produces two predictions: $\hat{k}^{t+1}$ and $\hat{f}^{t+1}$. We treat $\hat{k}^{t+1}$ with a Gaussian likelihood, while $\hat{f}^{t+1}$ is handled via a softmax-based likelihood.
The joint loss $\mathcal{L}$ is defined as below: 
$$\mathcal{L}^{t}=\frac{1}{\sigma_{1}^{2}}\mathcal{L}_{f}^{t}+\frac{1}{\sigma_{1}^{2}}L_{p}^{t}+log\sigma_{1}+log\sigma_{2},$$
where $\sigma_{1}$ and $\sigma_{2}$ as learning the relative weight of the losses $\mathcal{L}_{f}$ and $\mathcal{L}_{p}$ adaptively. 

\vspace{-1ex}
\subsection{Fine-tuning VLA with action labels}
Latent VLM that are pretrained to predict latent actions are not directly excutable on robots. Because latent actions are not actual end-effector actions or joint actions. To map latent actions to actual robot actions, we need to fine-tune latent VLM with an action head on a set of robot demonstrations datasets that contain ground truth labeled actions. 
We implement the action prediction module as a lightweight residual MLP attached to the decoder output. It takes the hidden representation from the last decoder layer as input and regresses real-valued actions, following the design pattern in \cite{kim2025fine}. Training is carried out by optimizing an L1 regression objective between predicted and ground-truth actions. In this stage , we adapting our model via LoRA fine-tuning\cite{hu2022lora}. 
\vspace{-1.5ex}

\section{EXPERIMENT}
\begin{table*}[t]
\centering
\caption{\textbf{Results on LIBERO benchmark across four evaluation
suites. } \Checkmark Methods do not utilize action labels during the pre-training stage. We reproduced results of LAPA and CoMo using the Prismatic-7B VLM and the same datasets on human, Bridge and RT-1. In the original CoMo paper, it is implemented based on the Diffusion Policy. $\dagger$Methods use additional wrist-view camera inputs. "w/o" indicates that no action labels are used during pre-training. Both methods for latent action are fine-tuned on RT-1 using action labels. 
}
\begin{tabular}{ccccccc}
\toprule
\textbf{Methods}&\textbf{w/o action label}&\textbf{Goal SR($\uparrow$))}&\textbf{Spatial SR($\uparrow$)}&\textbf{Object SR($\uparrow$))}&\textbf{Long SR($\uparrow$))}&\textbf{Avg. }\\
\midrule
OpenVLA\cite{kim2024openvla}&\XSolidBrush&76.5$\pm$1.0\% &\underline{82.6$\pm$0.9\%}&88.2$\pm$0.8\%&52.8$\pm$1.3\%&\underline{75.0}\\
Octo\cite{team2024octo}&\XSolidBrush&\underline{82.9$\pm$0.9\%}&76.1$\pm$1.0\%&84.3$\pm$0.9\%&51.6$\pm$1.8\%&73.7\\
Diffusion Policy\cite{chi2023diffusion}&\XSolidBrush&68.3$\pm$1.2\%&78.3$\pm$1.1\%&\underline{92.5$\pm$0.7\%}&50.5$\pm$2.4\%&72.4\\
MDT$^{\dagger}$\cite{reuss2024multimodal}&\XSolidBrush&71.2$\pm$0.8\%&77.3$\pm$1.0\%&88.1$\pm$1.2\%&\underline{62.6$\pm$1.4\%}&74.8\\
\midrule
LAPA\cite{ye2024latent}&\Checkmark&57.1$\pm$2.4\%&74.3$\pm$3.1\%&72.2$\pm$1.6\%&53.6$\pm$3.4\%&64.3\\
CoMo\cite{yang2025learning}&\Checkmark&63.2$\pm$1.2\%&76. 0$\pm$0.8\%&\underline{83.1$\pm$1.4\%}&54.6$\pm$1.8\%&69.2\\
\textbf{Ours} (human only)&\Checkmark&\underline{64.3$\pm$2.1\%}&70.2$\pm$1.9\%&69.5$\pm$2.4\%&49.2$\pm$3.5\%&63.3\\
\textbf{Ours} (Bridge only)&\Checkmark&67.5$\pm$1.7\%&75. 0$\pm$1.5\%&72.2$\pm$1.9\%&58.5$\pm$2.6\%&68.3\\
\textbf{Ours} (Bridge+RT-1)&\Checkmark&72. 0$\pm$1.6\%&80.5$\pm$2. 0\%&77.6$\pm$1.8\%&63.8$\pm$2. 0\%&73.5\\
\textbf{Ours (human+Bridge+RT-1)}&\Checkmark&\textbf{77.9$\pm$1.4\%}&\textbf{81.2$\pm$1.6\%}&\textbf{86.4$\pm$1.2\%}&\textbf{65.3$\pm$1.9\%}&\textbf{77.7}\\
\bottomrule
\end{tabular}
\label{tab:main result}
\vspace{-1.5ex}
    \caption{\textbf{The ablation experiments results on the LIBERO benchmark. }}
    \begin{tabular}{ccccccccc}
    \toprule
        \multirow{2}{*}{\centering Methods} & \multicolumn{2}{c}{Goal}&\multicolumn{2}{c}{Spatial}&\multicolumn{2}{c}{Object}&\multicolumn{2}{c}{Long} \\
        &LP-MSE($\downarrow$)&S-PCFC($\downarrow$)&LP-MSE($\downarrow$)&S-PCFC($\downarrow$)&LP-MSE($\downarrow$)&S-PCFC($\downarrow$)&LP-MSE($\downarrow$)&S-PCFC($\downarrow$)\\
        \midrule
         CoMo& 0.839&0.899	&0.881& \textbf{0.892}	&\textbf{0.662}& 0.902	&0.754&0.910\\
         LAPA&1.241&0.980&0.924&0.992&1.136&0.942&0.741&0.945\\
         Ours&\textbf{0.647}&\textbf{0.833}&\textbf{0.717}&0.945&0.690&\textbf{0.860}&\textbf{0.643}&\textbf{0.820}\\
        \bottomrule
    \end{tabular}
    \label{tab:ablation}
    \vspace{-3.5ex}
\end{table*}
In this section, we demonstrate the effectiveness of CARE as a latent action pretraining framework. Specifically, we focus on the following questions:

\noindent \textbf{Q1} (Performance): Can the CARE VLM trained without labels efficiently learn real actions during the fine-tuning stage to improve policy performance?

\noindent \textbf{Q2} (Interpretability): Do the continuous latent actions obtained by CARE have better action representation capabilities compared to latent actions obtained by other methods? 

\noindent \textbf{Q3} (Shortcut Learning): Does our method effectively prevent shortcut learning?
\vspace{-2ex}
\subsection{Setups}
\vspace{-1ex}
\noindent \textbf{Datasets.} As shown in Fig. \ref{fig:dataset}, we use two types of video datasets: robot and human manipulation videos. For robot data, we use 140k trajectories from the Open X-Embodiment \cite{o2024open}. For human videos, we use approximately 100k clips of daily activities from Something-Something v2 \cite{goyal2017something}. In total, our pretraining dataset comprises around 240k trajectories and video clips. For the fine-tuning stage, we use a 3\% uniform sample from the RT-1 dataset \cite{brohan2022rt}, which contains action labels. 

\noindent \textbf{Benchmark and Baselines.} We used LIBERO \cite{liu2023libero} as our benchmark. Our experiments were performed on four task suites: Spatial, Object, Goal, and Long. Each containing 10 tasks with 50 human-teleoperated demonstrations. We followed the OpenVLA \cite{kim2024openvla} data processing pipeline for all datasets. Our baselines include OpenVLA \cite{kim2024openvla}, Octo \cite{team2024octo}, DP \cite{chi2023diffusion}, MDT \cite{reuss2024multimodal}, LAPA \cite{ye2024latent}, and CoMo \cite{yang2025learning}. 

\noindent \textbf{Evaluation Metrics.} To address Q1, we evaluate baseline models and CARE in the LIBERO simulator using task success rate, where higher is better. 
For Q2, we conduct two experiments: (1) linear probing, following CoMo, where latent actions from CARE and CoMo are fed into an MLP to predict ground-truth actions, evaluated by LP-MSE (lower is better); (2) semantic label prediction on LAPA, CoMo, and CARE, evaluated by Semantic Accuracy (higher is better). 
For Q3, we adopt the S-PCFC metric \cite{yang2025learning}, where lower values indicate weaker shortcut learning. 
\vspace{-2ex}
\subsection{Main Results}
\vspace{-1ex}
The results presented in Tab. \ref{tab:main result} demonstrate that Our method CARE achieves higher success rates compared to other pre-training approaches without action labels, while remaining competitive with methods that utilize action labels for pre-training. Specifically, our method outperforms the action label-based pre-trained OpenVLA model by 1.4\% and 12.5\% in the Goal and Long tests, respectively. Moreover, compared to the label-free pre-training approaches of LAPA and CoMo based on autoregressive models, our method achieves superior performance across all four tasks. While slightly inferior to the diffusion policy-based CoMo approach in the Goal and Object tasks, our method demonstrates higher success rates in the Spatial and Long tasks. Furthermore, we observe that the success rate of the final experimental results improves as the pre-training data scales, validating that our method follows the scaling law. 
\vspace{-2ex}
\subsection{Ablation Study on Interpretability}
\vspace{-1ex}
To validate whether our pre-training method yields more interpretable latent action representations, we conducted linear probing experiments and semantic label prediction experiments, with results shown in Tab. \ref{tab:ablation} and Tab. \ref{tab:prediction} respectively. The results demonstrate that our method achieves lower LP-MSE scores than both CoMo and LAPA across Goal, Spatial, and Long tasks. Specifically:For the Goal task: 22.8\% reduction compared to CoMo and 47.8\% reduction compared to LAPA. For the Spatial task: 18.6\% reduction compared to CoMo and 22.4\% reduction compared to LAPA. For the Long task: 14.7\% reduction compared to CoMo and 13.2\% reduction compared to LAPA. As shown in Tab. \ref{tab:prediction}, the semantic label predictor trained solely on latent actions from the initial frame and subsequent 9 frames achieves 84.2\% accuracy on LIBERO Goal. This performance is comparable to that using 10 consecutive frames. These two experiments demonstrate that the latent actions obtained by our method exhibit strong semantic interpretability. 
\begin{table}[!htbp]
    \centering
    \vspace{-3.5ex}
    \caption{\textbf{Semantic label prediction. }}
    \begin{tabular}{cc}
    \toprule
        Input & Semantic Acc.($\uparrow$) \\
        \midrule
         Initial frame& 0.310\\
         Initial frame repeated by 10 times&0.326\\
         Initial frame + 9 subsequent frames&0.804\\
         Initial frame + 9 LAPA latent actions&0.641\\
         Initial frame + 9 CoMo latent actions&0.712\\
         Initial frame + 9 Ours latent actions&\textbf{0.842}\\
         \bottomrule
    \end{tabular}
    \label{tab:prediction}
    \vspace{-3ex}
\end{table}
\vspace{-1.5ex}
\subsection{Ablation Study on Shortcut Learning}
\vspace{-1ex}
Regarding S-PCFC, the results in Tab. \ref{tab:ablation} demonstrate that our method's latent actions effectively circumvent shortcut learning. Specifically, LAPA achieves near-perfect scores (0.992) across different tasks, indicating that VQ-VAE-based methods for extracting discrete latent actions remain susceptible to shortcut learning. The Q-former approach CoMo for extracting continuous latent actions partially mitigates this issue (0.892). Our method fundamentally addresses this through pretraining modification by leveraging VLM-based latent action extraction, it significantly avoids shortcut learning (0.833). 
\vspace{-2ex}
\subsection{Ablation Study on Multi-task Learning}
\vspace{-1ex}
To evaluate multi-task pre-training against single-task pre-training, we conducted ablation experiments with varied decoders while maintaining identical settings: multi-task, next-frame prediction, and keypoint trajectory prediction. As shown in Tab. \ref{tab:multi-task}, the multi-task Prismatic-7B model achieves a success rate (SR) of 77.7\% with a latent loss of 0.046, outperforming single-task models (67.8\% SR for frame prediction with 0.012 loss, and 54.3\% SR for point tracking with 0.009 loss). The lower SR of keypoint tracking supports prior findings favoring frame prediction, and these results confirm the synergistic benefits (1+1$>$2) of multi-task pre-training. 
\begin{table}[!htbp]
    \centering
    \vspace{-4ex}
      \caption{\textbf{Training details for different pre-training tasks.} SIMO denotes single-input multi-output, and latent loss refers to the task loss value at convergence during the experiment. }
    \begin{tabular}{cccccc}
    \toprule
        Model & Objective&SIMO&latent loss&SR \\
    \midrule
    \multirow{3}{*}{\centering Prismatic-7B}& frame predict&\XSolidBrush&0.012&67.8\\
    &point track&\XSolidBrush&0.009&54.3\\
    &multi-task&\Checkmark&0.046&77.7\\
    \bottomrule
    \end{tabular}
        \vspace{-4ex}
    \label{tab:multi-task}
\end{table}
\section{Conclusion}
\vspace{-2ex}
In this work, we introduce CARE, a multi-task pre-training approach that learns continuous latent action representations. By leveraging two decoders, we jointly pre-train a latent VLM. Compared to prior methods that rely on a separate latent action model, CARE achieves superior performance in simulations and produces latent actions that more effectively encode motions. Although a performance gap still exists compared to action-label-based pre-training, we plan to narrow this gap by incorporating action chunking, and multi-dimensional perceptual perspectives in future work. 
\newpage
\bibliographystyle{IEEEbib}
\bibliography{strings, refs}

@inproceedings{brohan2022rt,
  author       = {Anthony Brohan and
                  Noah Brown and
                  others},
  title        = {{RT-1:} Robotics Transformer for Real-World Control at Scale},
  booktitle    = {Robotics: Science and Systems XIX},
  year         = {2023},
  url          = {https://doi.org/10.15607/RSS.2023.XIX.025},
  doi          = {10.15607/RSS.2023.XIX.025},
  bibsource    = {dblp computer science bibliography, https://dblp.org}
}

@inproceedings{driess2023palm,
  author       = {Danny Driess and
                  Fei Xia and
                 others},
  title        = {PaLM-E: An Embodied Multimodal Language Model},
  booktitle    = {International Conference on Machine Learning},
  series       = {Proceedings of Machine Learning Research},
  volume       = {202},
  pages        = {8469--8488},
  year         = {2023},
  url          = {https://proceedings.mlr.press/v202/driess23a.html},
  bibsource    = {dblp computer science bibliography, https://dblp.org}
}

@article{bjorck2025gr00t,
  author       = {Johan Bjorck and
                  Fernando Casta{\~{n}}eda and
                  others},
  title        = {{GR00T} {N1:} An Open Foundation Model for Generalist Humanoid Robots},
  journal      = {CoRR},
  volume       = {abs/2503.14734},
  year         = {2025},
  url          = {https://doi.org/10.48550/arXiv.2503.14734},
  doi          = {10.48550/ARXIV.2503.14734},
  eprinttype    = {arXiv},
  eprint       = {2503.14734},
  bibsource    = {dblp computer science bibliography, https://dblp.org}
}

@inproceedings{o2024open,
  author       = {Abby O'Neill and
                  Abdul Rehman and
                  others},
  title        = {Open X-Embodiment: Robotic Learning Datasets and {RT-X} Models : Open
                  X-Embodiment Collaboration},
  booktitle    = {{IEEE} International Conference on Robotics and Automation},
  pages        = {6892--6903},
  year         = {2024},
  url          = {https://doi.org/10.1109/ICRA57147.2024.10611477},
  doi          = {10.1109/ICRA57147.2024.10611477},
  bibsource    = {dblp computer science bibliography, https://dblp.org}
}

@inproceedings{ye2024latent,
  author       = {Seonghyeon Ye and
                  Joel Jang and
                  others},
  title        = {Latent Action Pretraining from Videos},
  booktitle    = {The Thirteenth International Conference on Learning Representations},
  year         = {2025},
  url          = {https://openreview.net/forum?id=VYOe2eBQeh},
  bibsource    = {dblp computer science bibliography, https://dblp.org}
}

@inproceedings{schmidt2023learning,
  author       = {Dominik Schmidt and
                  Minqi Jiang},
  title        = {Learning to Act without Actions},
  booktitle    = {The Twelfth International Conference on Learning Representations},
  year         = {2024},
  url          = {https://openreview.net/forum?id=rvUq3cxpDF},
  bibsource    = {dblp computer science bibliography, https://dblp.org}
}

@InProceedings{chen2412moto,
    author    = {Chen, Yi and Ge, Yuying and others},
    title     = {Moto: Latent Motion Token as the Bridging Language for Learning Robot Manipulation from Videos},
    booktitle = {Proceedings of the IEEE/CVF International Conference on Computer Vision},
    year      = {2025},
    pages     = {19752-19763}
}

@inproceedings{karaev23cotracker,
  author       = {Nikita Karaev and
                  Ignacio Rocco and
                  others},
  title        = {CoTracker: It Is Better to Track Together},
  booktitle    = {Computer Vision - {ECCV} 2024 - 18th European Conference},
  series       = {Lecture Notes in Computer Science},
  volume       = {15120},
  pages        = {18--35},
  year         = {2024},
  url          = {https://doi.org/10.1007/978-3-031-73033-7\_2},
  doi          = {10.1007/978-3-031-73033-7\_2},
  bibsource    = {dblp computer science bibliography, https://dblp.org}
}

@article{yang2025learning,
  author       = {Jiange Yang and
                  Yansong Shi and
                  others},
  title        = {CoMo: Learning Continuous Latent Motion from Internet Videos for Scalable
                  Robot Learning},
  journal      = {CoRR},
  volume       = {abs/2505.17006},
  year         = {2025},
  url          = {https://doi.org/10.48550/arXiv.2505.17006},
  doi          = {10.48550/ARXIV.2505.17006},
  eprinttype    = {arXiv},
  eprint       = {2505.17006},
  bibsource    = {dblp computer science bibliography, https://dblp.org}
}

@inproceedings{van2017neural,
  author       = {A{\"{a}}ron van den Oord and
                  Oriol Vinyals and
                  Koray Kavukcuoglu},
  title        = {Neural Discrete Representation Learning},
  booktitle    = {Advances in Neural Information Processing Systems 30: Annual Conference
                  on Neural Information Processing Systems},
  pages        = {6306--6315},
  year         = {2017},
  url          = {https://proceedings.neurips.cc/paper/2017/hash/7a98af17e63a0ac09ce2e96d03992fbc-Abstract.html},
  bibsource    = {dblp computer science bibliography, https://dblp.org}
}

@inproceedings{bruce2024genie,
  author       = {Jake Bruce and
                  Michael D. Dennis and
                  others},
  title        = {Genie: Generative Interactive Environments},
  booktitle    = {Forty-first International Conference on Machine Learning},
  year         = {2024},
  url          = {https://openreview.net/forum?id=bJbSbJskOS},
  bibsource    = {dblp computer science bibliography, https://dblp.org}
}

@article{mccarthy2024towards,
  author       = {Robert McCarthy and
                  Daniel Chee Hian Tan and
                  others},
  title        = {Towards Generalist Robot Learning from Internet Video: {A} Survey},
  journal      = {J. Artif. Intell. Res.},
  volume       = {83},
  year         = {2025},
  url          = {https://doi.org/10.1613/jair.1.17400},
  doi          = {10.1613/JAIR.1.17400},
  bibsource    = {dblp computer science bibliography, https://dblp.org}
}

@inproceedings{karamcheti2024prismatic,
  author       = {Siddharth Karamcheti and
                  Suraj Nair and
                  others},
  title        = {Prismatic VLMs: Investigating the Design Space of Visually-Conditioned
                  Language Models},
  booktitle    = {Forty-first International Conference on Machine Learning},
  year         = {2024},
  url          = {https://openreview.net/forum?id=6FXtu8clyp},
  bibsource    = {dblp computer science bibliography, https://dblp.org}
}

@inproceedings{zhai2023sigmoid,
  author       = {Xiaohua Zhai and
                  Basil Mustafa and
                  others},
  title        = {Sigmoid Loss for Language Image Pre-Training},
  booktitle    = {{IEEE/CVF} International Conference on Computer Vision},
  pages        = {11941--11952},
  year         = {2023},
  url          = {https://doi.org/10.1109/ICCV51070.2023.01100},
  doi          = {10.1109/ICCV51070.2023.01100},
  bibsource    = {dblp computer science bibliography, https://dblp.org}
}

@article{oquab2023dinov2,
  author       = {Maxime Oquab and
                  Timoth{\'{e}}e Darcet and
                  others},
  title        = {DINOv2: Learning Robust Visual Features without Supervision},
  journal      = {Trans. Mach. Learn. Res.},
  year         = {2024},
  url          = {https://openreview.net/forum?id=a68SUt6zFt},
  bibsource    = {dblp computer science bibliography, https://dblp.org}
}

@inproceedings{kim2024openvla,
  author       = {Moo Jin Kim and
                  Karl Pertsch and
                  others},
  title        = {OpenVLA: An Open-Source Vision-Language-Action Model},
  booktitle    = {Conference on Robot Learning},
  series       = {Proceedings of Machine Learning Research},
  volume       = {270},
  pages        = {2679--2713},
  year         = {2024},
  url          = {https://proceedings.mlr.press/v270/kim25c.html},
  bibsource    = {dblp computer science bibliography, https://dblp.org}
}

@inproceedings{kendall2018multi,
  author       = {Alex Kendall and
                  Yarin Gal and
                  Roberto Cipolla},
  title        = {Multi-Task Learning Using Uncertainty to Weigh Losses for Scene Geometry
                  and Semantics},
  booktitle    = {2018 {IEEE} Conference on Computer Vision and Pattern Recognition},
  pages        = {7482--7491},
  year         = {2018},
  url          = {http://openaccess.thecvf.com/content\_cvpr\_2018/html/Kendall\_Multi-Task\_Learning\_Using\_CVPR\_2018\_paper.html},
  doi          = {10.1109/CVPR.2018.00781},
  bibsource    = {dblp computer science bibliography, https://dblp.org}
}

@article{kim2025fine,
  author       = {Moo Jin Kim and
                  Chelsea Finn and
                  Percy Liang},
  title        = {Fine-Tuning Vision-Language-Action Models: Optimizing Speed and Success},
  journal      = {CoRR},
  volume       = {abs/2502.19645},
  year         = {2025},
  url          = {https://doi.org/10.48550/arXiv.2502.19645},
  doi          = {10.48550/ARXIV.2502.19645},
  eprinttype    = {arXiv},
  eprint       = {2502.19645},
  bibsource    = {dblp computer science bibliography, https://dblp.org}
}

@inproceedings{hu2022lora,
  author       = {Yuhui Xu and
                  Lingxi Xie and
                  others},
  title        = {QA-LoRA: Quantization-Aware Low-Rank Adaptation of Large Language
                  Models},
  booktitle    = {The Twelfth International Conference on Learning Representations},
  year         = {2024},
  url          = {https://openreview.net/forum?id=WvFoJccpo8},
  bibsource    = {dblp computer science bibliography, https://dblp.org}
}

@inproceedings{liu2023libero,
  author       = {Bo Liu and
                  Yifeng Zhu and
                  others},
  title        = {{LIBERO:} Benchmarking Knowledge Transfer for Lifelong Robot Learning},
  booktitle    = {Advances in Neural Information Processing Systems 36: Annual Conference
                  on Neural Information Processing Systems},
  year         = {2023},
  url          = {http://papers.nips.cc/paper\_files/paper/2023/hash/8c3c666820ea055a77726d66fc7d447f-Abstract-Datasets\_and\_Benchmarks.html},
  bibsource    = {dblp computer science bibliography, https://dblp.org}
}

@inproceedings{goyal2017something,
  author       = {Raghav Goyal and
                  Samira Ebrahimi Kahou and
                  others},
  title        = {The "Something Something" Video Database for Learning and Evaluating
                  Visual Common Sense},
  booktitle    = {{IEEE} International Conference on Computer Vision},
  pages        = {5843--5851},
  year         = {2017},
  url          = {https://doi.org/10.1109/ICCV.2017.622},
  doi          = {10.1109/ICCV.2017.622},
  bibsource    = {dblp computer science bibliography, https://dblp.org}
}

@inproceedings{team2024octo,
  author       = {Dibya Ghosh and
                  Homer Rich Walke and
                  others},
  title        = {Octo: An Open-Source Generalist Robot Policy},
  booktitle    = {Robotics: Science and Systems XX},
  year         = {2024},
  url          = {https://doi.org/10.15607/RSS.2024.XX.090},
  doi          = {10.15607/RSS.2024.XX.090},
  bibsource    = {dblp computer science bibliography, https://dblp.org}
}

@inproceedings{reuss2024multimodal,
  author       = {Moritz Reuss and
                  {\"{O}}mer Erdin{\c{c}} Yagmurlu and
                  Fabian Wenzel and
                  Rudolf Lioutikov},
  title        = {Multimodal Diffusion Transformer: Learning Versatile Behavior from
                  Multimodal Goals},
  booktitle    = {Robotics: Science and Systems XX},
  year         = {2024}
}

@inproceedings{chi2023diffusion,
  author       = {Cheng Chi and
                  Siyuan Feng and
                  others},
  title        = {Diffusion Policy: Visuomotor Policy Learning via Action Diffusion},
  booktitle    = {Robotics: Science and Systems XIX},
  year         = {2023}
}

@inproceedings{nav,
    title = "{RATE}-Nav: Region-Aware Termination Enhancement for Zero-shot Object Navigation with Vision-Language Models",
    author = "Li, Junjie  and
      Zhang, Nan  and
      Qu, Xiaoyang  and
      Lu, Kai  and
      Li, Guokuan  and
      Wan, Jiguang  and
      Wang, Jianzong",
    booktitle = "Findings of the Association for Computational Linguistics: ACL 2025",
    year = "2025",
    pages = "6564--6574",
}

@inproceedings{jiaziqi,
    title = "Hierarchical-Task-Aware Multi-modal Mixture of Incremental {L}o{RA} Experts for Embodied Continual Learning",
    author = "Jia, Ziqi  and
      Wang, Anmin  and
      Qu, Xiaoyang  and
      Yang, Xiaowen  and
      Wang, Jianzong",
    booktitle = "Proceedings of the 63rd Annual Meeting of the Association for Computational Linguistics (Volume 1: Long Papers)",
    year = "2025",
    pages = "28415--28427",
}

@INPROCEEDINGS{jiaziqi2,
  author={Jia, Ziqi and Li, Junjie and Qu, Xiaoyang and Wang, Jianzong},
  booktitle={2025 IEEE International Conference on Robotics and Automation (ICRA)}, 
  title={Enhancing Multi-Agent Systems via Reinforcement Learning with LLM-Based Planner and Graph-Based Policy}, 
  year={2025},
  pages={1240-1246},
  }

@INPROCEEDINGS{pointaction,
  author={Tao, Wei and He, Shenglin and Qu, Xiaoyang and Wan, Jiguang and Wang, Jianzong},
  booktitle={ICASSP 2025 - 2025 IEEE International Conference on Acoustics, Speech and Signal Processing (ICASSP)}, 
  title={PointActionCLIP: Preventing Transfer Degradation in Point Cloud Action Recognition with a Triple-Path CLIP}, 
  year={2025},
  pages={1-5},
}

@inproceedings{aaai,
author = {Zhang, Bin and Chen, Jinggang and Qu, Xiaoyang and Li, Guokuan and Lu, Kai and Wan, Jiguang and Xiao, Jing and Wang, Jianzong},
title = {RUNA: object-level out-of-distribution detection via regional uncertainty alignment of multimodal representations},
year = {2025},
booktitle = {Proceedings of the Thirty-Ninth AAAI Conference on Artificial Intelligence and Thirty-Seventh Conference on Innovative Applications of Artificial Intelligence and Fifteenth Symposium on Educational Advances in Artificial Intelligence},
articleno = {2943},
numpages = {9},
}

\end{document}